\documentclass[final,5p,times,twocolumn]{elsarticle}

\usepackage{amssymb}
\usepackage{longtable} 
\usepackage{pdflscape} 
\usepackage{rotating}

\usepackage[T1]{fontenc} 
\usepackage[utf8]{inputenc}
\usepackage[english]{babel}

\usepackage{longtable} 
\usepackage{pdflscape} 
\usepackage{rotating}
\usepackage{blindtext}
\usepackage{hyperref}
\usepackage{multirow}
\usepackage{soul} 

\hypersetup{
    colorlinks=true,
    linkcolor=blue,
    filecolor=magenta,      
    urlcolor=cyan
    }

\journal{Elsevier Journal}

\begin{document}

\begin{frontmatter}

\title{Exploring the Relationships Between Physiological Signals During Automated Fatigue Detection}
\author[inst1]{Kourosh Kakhi}
\affiliation[inst1]{organization={Institute for Intelligent Systems Research and Innovation (IISRI)}, 
            city={Geelong},
            postcode={3217}, 
            state={VIC},
            country={Australia}}

\affiliation[inst2]{organization={School of Mathematics, Physics and Computing, University of Southern Queensland}, 
            city={Springfield},
            postcode={}, 
            state={QLD},
            country={Australia}}

\author[inst1]{Abbas Khosravi}
\author[inst1]{Roohallah Alizadehsani}
\author[inst2]{U. Rajendra Acharya}
\author[inst3]{}

\begin{abstract}
\textbf{Background:} Fatigue detection through physiological signals has gained growing relevance across safety-critical domains such as transportation, healthcare, and human performance monitoring. While many studies focus on individual modalities (e.g., EEG or ECG), limited attention has been given to investigating statistical relationships between signal pairs as a means to enhance classification robustness. This study aims to explore how inter-signal statistical features correlation, cross-correlation, and covariance across multiple physiological signals can support fatigue state prediction.

\textbf{Methodology:} Using the DROZY dataset, we extracted pairwise statistical features from four physiological signals: ECG, EMG, EOG, and EEG. Fifteen distinct signal combinations were evaluated, covering uni-modal to multi-modal configurations. Feature extraction emphasized statistical relationships between signals rather than raw amplitude characteristics. The extracted features were fed into four supervised machine learning classifiers: Decision Tree (DT), Random Forest (RF), Logistic Regression (LR), and XGBoost (XGB). Performance was assessed using accuracy, precision, recall, and area under the curve (AUC). Additionally, SHAP (SHapley Additive exPlanations) values were computed to evaluate feature importance and interpret model behavior.

\textbf{Results:} Among all classifiers and signal combinations, XGBoost applied to the EMG\,|\,EEG combination achieved the highest classification performance, with an accuracy of 0.888 and an AUC of 0.975. SHAP-based ranking revealed that the correlation between ECG and EOG-H was the most influential feature across models. Feature interaction plots indicated non-linear relationships between statistical measures and fatigue levels. The multi-signal approach consistently outperformed single-signal models, with combinations involving EEG and EMG contributing most significantly to predictive power.

\textbf{Conclusion:} This study demonstrates that analyzing statistical relationships between physiological signals offers substantial improvements in fatigue classification performance. Signal combinations, particularly EMG and EEG, provide complementary information that enhances model accuracy and robustness. SHAP analysis highlighted the interpretability of feature contributions, reinforcing the utility of statistical fusion approaches in physiological signal processing. These findings support the integration of multi-signal, feature-level fusion strategies in practical fatigue monitoring systems using wearable technology.

\end{abstract}




\end{frontmatter}


\section{Introduction}\label{sec:intro}
 Fatigue is a progressive and pervasive psychophysiological state marked by diminished alertness, impaired motor coordination, slowed cognitive processing, and a decline in task performance. Its presence is particularly critical in high-stakes occupational domains such as healthcare, aviation, transportation, and manufacturing where momentary lapses in attention can result in catastrophic consequences \cite{1}. Extensive research has demonstrated the detrimental effects of fatigue on reaction time, decision-making, and situational awareness, emphasizing the need for effective, real-time monitoring systems capable of identifying fatigue onset before performance deterioration manifests \cite{2}. Despite its significance, fatigue remains a challenging construct to quantify objectively due to its multifactorial nature and variability across individuals and contexts.

In recent years, physiological signal analysis has emerged as a promising avenue for fatigue detection, offering objective and continuous assessment capabilities \cite{3}. Unlike traditional behavioral approaches which rely on observable features such as eye-blink frequency, facial microexpressions, or driving patterns physiological signals provide a direct window into the body’s internal regulatory mechanisms. Biosignals such as the electroencephalogram (EEG) \cite{4}, electrocardiogram (ECG) \cite{5}, electromyogram (EMG) \cite{6}, and electrooculogram (EOG) \cite{7} are particularly valuable in this regard, as they encode real-time information related to central nervous system activity, autonomic regulation, and muscular engagement each of which is sensitive to fatigue induced alterations. The integration of multiple physiological modalities has shown considerable potential for improving the robustness and generalizability of fatigue detection systems, especially when fused at the feature or decision level \cite{8}.

While prior research has primarily focused on signal specific features derived from time, frequency, or nonlinear domains, there is growing recognition that inter-signal relationships may contain latent information crucial for fatigue classification. Recent advances in machine learning and explainable artificial intelligence (XAI) have further enabled the modeling of complex, high-dimensional physiological datasets, allowing researchers to uncover subtle patterns that may not be accessible through conventional statistical analysis \cite{9}. In our previous works, we explored the utility of SHAP-based feature attribution methods for interpreting fatigue models trained on single- modality and multi-modality features. Furthermore, we evaluated the contribution of signal characteristics such as amplitude and distributional behavior across different machine learning classifiers including Random Forest (RF) \cite{10}, Logistic Regression (LR) \cite{11} , Decision Trees (DT) \cite{12}, and Extreme Gradient Boosting (XGBoost) \cite{13}.

Building upon this foundation, the current study presents a comprehensive investigation into the statistical interdependencies between physiological signal pairs and their efficacy in fatigue state classification. Rather than relying on raw signal values or standard time-series descriptors, we introduce a feature set that captures pairwise statistical relationships namely correlation, cross-correlation, and covariance between signals. These second order descriptors provide insight into how physiological systems co-regulate during fatigue progression and allow for the detection of synchronized changes across modalities. The motivation stems from the hypothesis that fatigue does not affect physiological signals in isolation but manifests through complex, interdependent patterns observable when signal pairs are analyzed jointly \cite{14}.

To empirically validate this hypothesis, we utilize the DROZY dataset, a publicly available multimodal sleep and fatigue dataset containing labeled recordings of EEG, ECG, EMG, and EOG signals. We construct 15 signal combinations, encompassing unimodal, bimodal, trimodal, and quad-modal pairings, and extract pairwise statistical features across each windowed segment. These features are then input into four supervised machine learning classifiers DT, RF, LR, and XGBoost to evaluate their ability to distinguish fatigue levels based on the Karolinska Sleepiness Scale (KSS). Evaluation metrics include accuracy, precision, recall, and area under the receiver operating characteristic curve (AUC). To further interpret model behavior and validate the contribution of each feature, we employ SHAP analysis and feature importance scores derived from ensemble models.

We propose a structured methodology using machine learning models to:
\begin{itemize}
    \item Investigate pairwise combinations of physiological signals based on statistical descriptors.
    
    \item Evaluate model performance in terms of classification metrics.

    \item Interpret results using explainable AI techniques to identify signal pairings with the highest predictive contribution.

\end{itemize}
By emphasizing the statistical alignment and distributional behavior of paired signals, this work provides a fresh perspective on physiological data fusion for fatigue detection. The results offer actionable insights for developing more reliable and generalizable fatigue monitoring systems and highlight the nuanced roles that different bio-signals play when analyzed through statistical lenses. Future directions include expanding the scope to include real-time applications and wearable device integration for continuous fatigue monitoring.
\section{Data}\label{sec:data}
This study utilizes the publicly available DROZY dataset, a benchmark physiological database designed for sleepiness and fatigue related research \cite{15}. Collected under controlled experimental conditions, DROZY provides a rich, multimodal record of human physiological responses during fatigue inducing protocols. The dataset includes synchronized recordings of electroencephalogram (EEG), electrooculogram (EOG), electromyogram (EMG), and electrocardiogram (ECG) signals from 14 healthy participants subjected to cognitively demanding tasks over extended periods, intended to induce varying degrees of mental fatigue.

Each participant underwent sessions involving sustained attention tasks, during which physiological signals were continuously recorded using clinical-grade wearable equipment. The recordings span over 1,100 labeled epochs, with each 30 second segment annotated using the Karolinska Sleepiness Scale (KSS) \cite{16}, a validated subjective fatigue index ranging from 1 (extremely alert) to 9 (very sleepy, fighting sleep). In this study, we grouped KSS ratings into three fatigue levels low (1--3), medium (4--6), and high (7--9) to facilitate multi-class classification modeling.

The DROZY dataset offers a unique opportunity for pairwise signal analysis due to its high temporal resolution and multi-channel structure. Specifically, the EEG channels (C3, C4, Cz, Fz, Pz), EOG (horizontal and vertical components), ECG, and surface EMG provide coverage of both central and peripheral nervous system dynamics. To ensure consistency in signal interpretation, all signals were resampled and segmented into overlapping windows, allowing for temporal alignment across modalities. Preprocessing steps included band-pass filtering, artifact rejection, and z-score normalization, ensuring the robustness of subsequent statistical analysis.

Unlike prior studies that focus on intra-signal characteristics (e.g., power spectra or entropy) \cite{17} \cite{18} \cite{19}, our work leverages the DROZY dataset to explore inter-signal statistical relationships. This is achieved by extracting second order descriptors Pearson correlation, cross-correlation, and covariance between signal pairs across the synchronized epochs. A total of 15 unique signal combinations were constructed, covering unimodal (e.g., EEG--EEG), bimodal (e.g., ECG--EOG), trimodal, and fully fused (ECG--EOG--EMG--EEG) groupings. These pairings allow us to probe how synchronous variation across modalities reflects the onset and progression of fatigue.

The resulting feature matrix comprised over 110 extracted statistical features per segment, capturing spatial and temporal dependencies across physiological modalities. Each segment was labeled based on grouped KSS scores representing low, medium, or high fatigue levels. The dataset was stratified by class labels and randomly split into 70\% training and 30\% testing subsets. While class distribution was preserved, future work may incorporate participant-level separation to further reduce inter-subject bias and improve generalizability.

By exploiting the multimodal richness of the DROZY dataset through pairwise statistical descriptors, this study offers a novel perspective on how interdependencies between physiological signals evolve with fatigue. The dataset’s granularity, multi-class labels, and synchronized sensor streams make it a strong foundation for evaluating signal alignment as a discriminative factor in real-world fatigue detection systems.


\section{Methodology}\label{sec:meth}

\begin{figure}[h]
    \centering
    \includegraphics[scale=0.4]{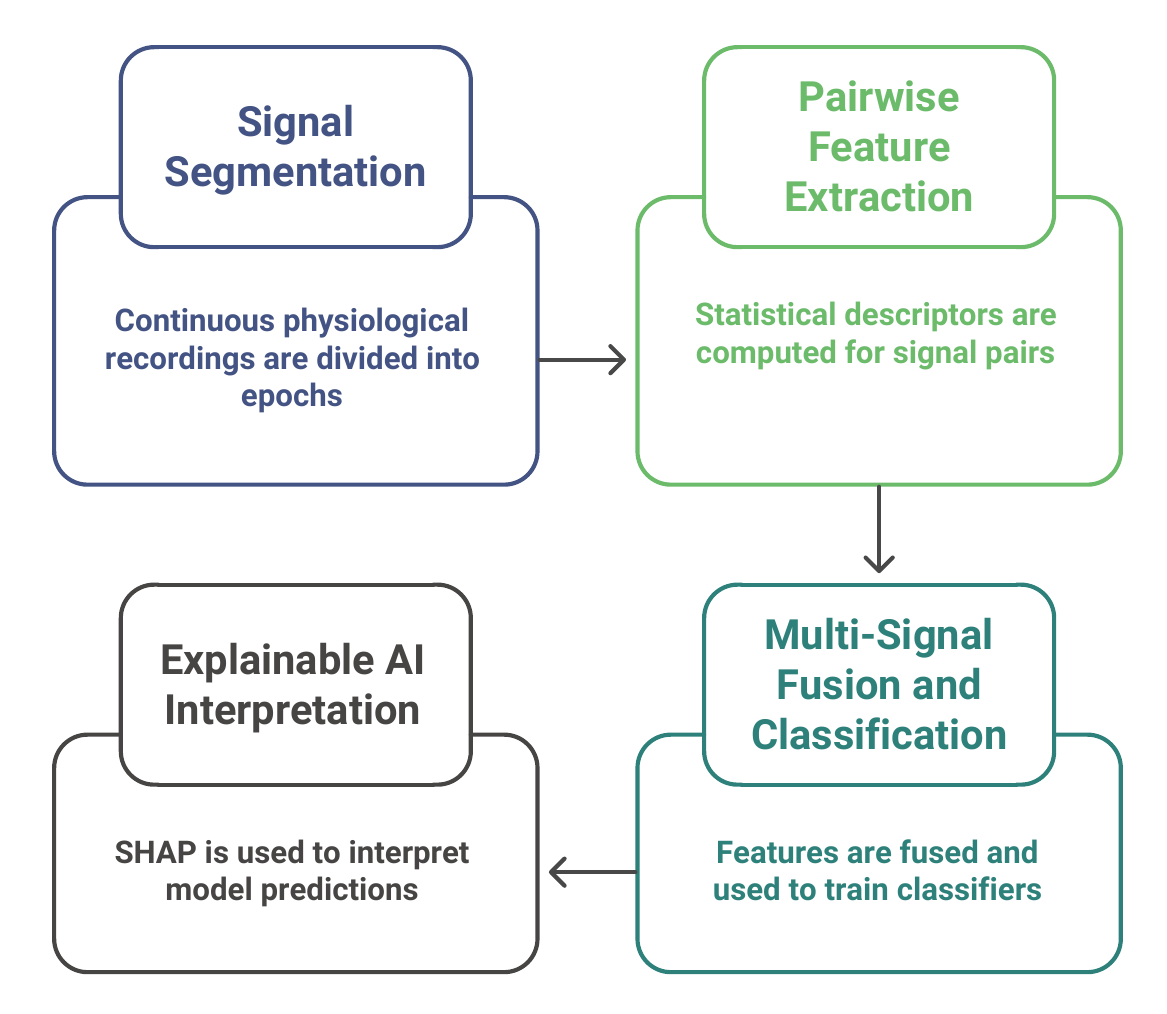}
    \caption{Overview of the proposed fatigue detection pipeline.}
    \label{fig:Pipeline}
\end{figure}

This study investigates the statistical interdependencies between pairs of physiological signals to improve the accuracy and robustness of fatigue detection models (Figure \ref{fig:Pipeline}). While prior research has primarily focused on analyzing individual signals in isolation (e.g., EEG or ECG) \cite{20} \cite{21}, our approach emphasizes the importance of cross-modal interactions, particularly the temporal coordination and statistical co-variation between signals such as electrocardiography (ECG), electroencephalography (EEG), electromyography (EMG), and electrooculography (EOG). The underlying hypothesis is that fatigue-related physiological alterations may not be fully captured by uni-variate descriptors alone but are more reliably detected by analyzing how these signals interact and evolve together over time.
To explore this hypothesis, we established a structured analytical pipeline comprising four main stages:
\begin{enumerate}
    \item Signal Segmentation: Continuous physiological recordings were segmented into fixed-duration epochs to capture transient fluctuations associated with fatigue dynamics \cite{22}.
    \item Pairwise Feature Extraction: For each signal pair, we computed statistical descriptors capturing correlation, cross-correlation, and covariance across signal windows, providing insights into temporal synchrony and joint variability \cite{23}.
    \item Multi-Signal Fusion and Classification: Extracted features were fused and used to train supervised classifiers, enabling the prediction of fatigue states as labeled by the Karolinska Sleepiness Scale (KSS) \cite{24}.
    \item Explainable AI Interpretation: We employed SHAP (SHapley Additive exPlanations) to attribute model predictions to specific signal combinations and feature interactions, offering transparency into which physiological relationships contributed most to classification decisions \cite{25}.
\end{enumerate}
\begin{figure*}[t]
    \centering
    \includegraphics[scale=0.8]{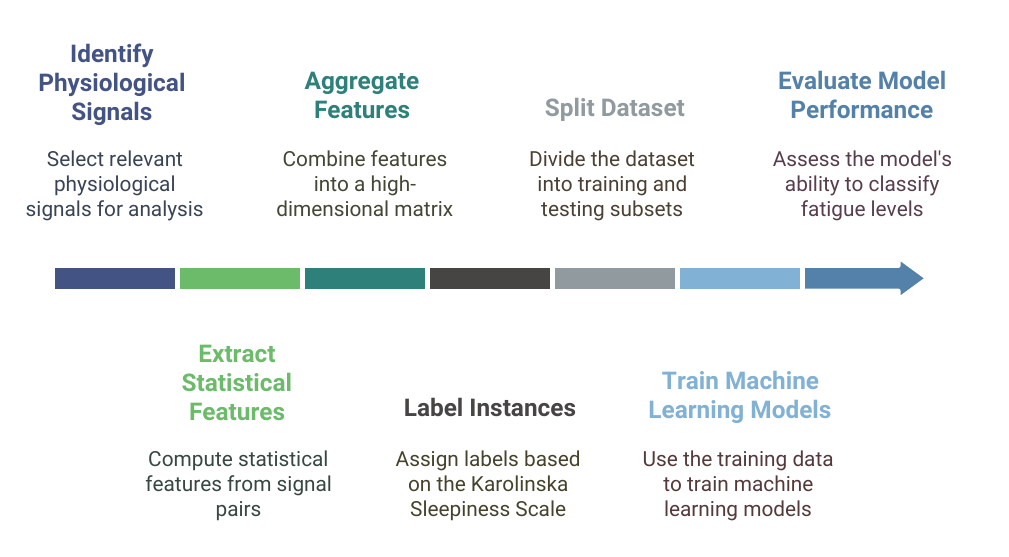}
    \caption{Multi-Signal Fusion and Dataset Construction.}
    \label{fig:Multi-Signal Fusion and Dataset Construction}
\end{figure*}


\subsection{Signal Pre-processing and Segmentation}\label{sec:Signal Preprocessing and Segmentation}
All physiological recordings used in this study were obtained from the publicly available DROZY dataset, a comprehensive multimodal physiological database designed to support sleepiness and fatigue detection research. The dataset includes high-resolution, synchronized recordings of electroencephalography (EEG), electrooculography (EOG), electromyography (EMG), and electrocardiography (ECG) collected from healthy participants during extended wakefulness and cognitive task performance designed to elicit fatigue.
\subsubsection{Signal Alignment and Filtering}\label{Subsec:Signal Alignment and Filtering}
To ensure uniformity across modalities, all signals were resampled to a common sampling rate where necessary, and signal alignment was verified based on timestamp synchronization provided in the raw EDF files \cite{26}. Each signal underwent standard preprocessing steps to enhance signal quality and suppress noise. This included:

\begin{itemize}
    \item Baseline correction to remove DC drift and low-frequency movement artifacts.
    \item Bandpass filtering (typically between 0.5 Hz and 45 Hz) to isolate relevant physiological frequency bands while attenuating muscle noise, power-line interference, and signal saturation artifacts.
    \item Z-score normalization applied per signal channel to standardize across participants and sessions, facilitating inter-subject comparability.
\end{itemize}

\subsubsection{Temporal Segmentation}\label{Subsec:Temporal Segmentation}
Following preprocessing, each continuous signal stream was segmented into fixed-duration, non-overlapping windows of 20 seconds (corresponding to 10,240 samples at 512 Hz). This window length was selected to strike a balance between capturing meaningful temporal dynamics and ensuring adequate statistical stability for feature computation. Each segment served as an independent instance for feature extraction and classification.

\subsubsection{Labeling and Fatigue Categorization}\label{Subsec:Labeling and Fatigue Categorization}
Each 20-second window was assigned a corresponding Karolinska Sleepiness Scale (KSS) label, a subjective self-report metric scored on a 9-point ordinal scale. In line with common practice and your previous studies, these KSS values were collapsed into three fatigue categories to enable multiclass classification:

\begin{itemize}
    \item Alert (KSS 1–3): No fatigue or drowsiness reported.
    \item Normal (KSS 4–6): Mild to moderate fatigue.
    \item Fatigued (KSS 7–9): High fatigue, drowsiness, or microsleep onset.
\end{itemize}
These categories were encoded as integer class labels (0: alert, 1: normal, 2: fatigued), forming the basis for supervised learning.

\subsubsection{Class Distribution and Validation Considerations}\label{Subsec:Class Distribution and Validation Considerations}

To ensure model robustness and avoid overfitting participant-specific patterns, class distributions were monitored throughout the segmentation process. The data was later split into stratified training and testing sets, preserving the balance of the class across folds. Where possible, participant independence was preserved between folds to reduce subject leakage and improve generalizability among unseen individuals.

\subsection{Feature Extraction from Pairwise Signal Combinations}\label{sec:Feature Extraction from Pairwise Signal Combinations}
For each segment, we extracted statistical descriptors between all unique pairwise combinations of signals. These features capture the co-behavior of two physiological systems and quantify their alignment, fluctuation coherence, and amplitude similarity. The extracted features fall into three main categories:
\subsubsection{Correlation Features}\label{Subsec:Correlation Features}
Pearson's correlation coefficient ($r$) \cite{27} was computed for each pair of signals to capture the linear similarity between their temporal fluctuations:

\[
r = \frac{\sum_{i=1}^{N}(x_i - \bar{x})(y_i - \bar{y})}
         {\sqrt{\sum_{i=1}^{N}(x_i - \bar{x})^2} \cdot 
          \sqrt{\sum_{i=1}^{N}(y_i - \bar{y})^2}}
\]

where \( x_i \) and \( y_i \) are the values of the two signals at time index \( i \), 
\( \bar{x} \) and \( \bar{y} \) are their means, and \( N \) is the number of time points in the window.

\subsubsection{Covariance Features}\label{Subsec:Covariance Features}
The covariance \( \mathrm{Cov}(X, Y) \) \cite{28} was computed to quantify the joint variability of two signals, 
indicating their synchronous amplitude shifts:

\[
\mathrm{Cov}(X, Y) = \frac{1}{N} \sum_{i=1}^{N} (x_i - \bar{x})(y_i - \bar{y})
\]

This metric is sensitive to both magnitude and direction of co-fluctuation and provides complementary information to the correlation coefficient.

\subsubsection{Cross-Correlation Features}\label{Subsec:Cross-Correlation Features}
To capture potential time-lagged dependencies and temporal alignment, we computed the maximum cross-correlation coefficient \cite{29} within a \( \pm 500 \, \mathrm{ms} \) window:

\[
\rho_{xy}(\tau) = \max_{\tau} \left[ \frac{1}{N} \sum_{i=1}^{N} x_i \cdot y_{i+\tau} \right]
\]
Here, \( \tau \) is the lag in time steps, and \( y_{i+\tau} \) denotes the shifted version of signal \( y \). 
This descriptor highlights interactions where one physiological response may follow another with delay 
(e.g., heart rate following neural activation).

\subsection{Multi-Signal Fusion and Dataset Construction}\label{sec:Multi-Signal Fusion and Dataset Construction}
Each pair of signals generated a feature vector composed of their correlation, covariance, and cross-correlation descriptors. These vectors were aggregated into a high-dimensional feature matrix comprising over 117 pairwise statistical features per segment (Table \ref{tab:FeatureExtraction}). Each instance was labeled based on KSS grouping and stored for classifier training. The dataset was then split into training (70\%) and testing (30\%) subsets, ensuring participant-level separation to avoid subject-specific bias and overfitting.
\begin{table}[h]
    \centering
    \caption{Total Features Extracted.}
    \label{tab:FeatureExtraction}
    \begin{tabular}{l c} 
        \hline
        \textbf{Feature Type} & \textbf{Total Features} \\ \hline
        Correlation & 41 \\
        Cross-Correlation & 40 \\
        Covariance & 36 \\
        \hline
        Total Features & 117 \\ 
        \hline
    \end{tabular}
\end{table}

\subsection{Classifier Training and Evaluation}\label{sec:Classifier Training and Evaluation}
Four widely-used supervised learning models were employed to classify fatigue levels:
\begin{itemize}
    \item Decision Tree (DT)
    \item Random Forest (RF)
    \item Logistic Regression (LR)
    \item Extreme Gradient Boosting (XGBoost)    
\end{itemize}
Each model was evaluated using accuracy, precision, recall, and AUC. XGBoost consistently yielded the highest performance, with the top-performing signal combination (EMG–EEG) achieving 88.81\% accuracy and 0.9747 AUC.

\subsection{Explainability via SHAP Analysis}\label{sec:Explainability via SHAP Analysis}
To provide interpretability and identify the most informative signal pair features, the SHAP values (SHapley Additive exPlanations) were computed for each trained model. These values quantify the contribution of each feature to a specific prediction, grounded in cooperative game theory. The mean absolute SHAP values were used to rank global feature importance, while dependence plots and interaction plots revealed nonlinear interactions and feature synergy. For example, the feature \textit{ECG\_EOG-H\_Corr} consistently emerged as a dominant predictor in the models.

Figure \ref{fig:Multi-Signal Fusion and Dataset Construction} presents a structured methodological pipeline for exploring statistical interdependencies between physiological signals in the context of fatigue detection. By segmenting synchronized multimodal recordings into fixed temporal windows and extracting second-order statistical features specifically correlation, cross-correlation, and covariance between all possible signal pairs, we created a rich and interpretable feature space. This was followed by classification using multiple machine learning algorithms, with rigorous evaluation metrics to assess performance across different fusion configurations. The integration of explainability techniques, such as SHAP value analysis, further improved the interpretability of model outputs by attributing decision contributions to specific signal interactions. This methodological framework enables the identification of signal combinations that are most indicative of fatigue and provides a foundation for developing robust, interpretable, and generalizable fatigue detection systems based on physiological data.

\section{Machine Learning Models}\label{sec:ML Learning}
To rigorously evaluate the predictive potential of statistical features derived from physiological signal pairings, this study employed four established supervised machine learning algorithms: Decision Tree (DT), Random Forest (RF), Logistic Regression (LR), and Extreme Gradient Boosting (XGBoost). The selected models span a spectrum of complexity and learning paradigms ranging from interpretable decision-based structures to powerful ensemble techniques enabling a multi-faceted analysis of fatigue classification performance \cite{30}.

Each classifier was trained on a unified feature matrix constructed from pairwise statistical descriptors namely Pearson correlation, cross correlation (with a ±500 ms lag), and covariance extracted from synchronized 20-second non-overlapping signal segments. The corresponding class labels were derived from discretized Karolinska Sleepiness Scale (KSS) scores, reflecting distinct fatigue states. Prior to model training, standardization was applied where appropriate, particularly for LR, to ensure feature comparability and optimization stability.

The Decision Tree classifier was implemented using the Gini impurity criterion, enabling straightforward interpretation of hierarchical decision logic based on the most discriminative features. The Random Forest model, comprising 1000 estimators, enhanced classification robustness through bagging and random subspace selection, thereby mitigating overfitting and reducing model variance. Logistic Regression, serving as a linear probabilistic baseline, estimated class likelihoods as a function of linear combinations of input features. XGBoost, a highly regularized gradient boosting framework, was adopted to capture intricate non-linear dependencies, benefiting from shrinkage, column sampling, and iterative residual fitting to achieve state-of-the-art predictive performance.

All models were evaluated using a subject wise stratified 70/30 train-test split to ensure participant level independence and prevent model overfitting to individual specific characteristics. Performance assessment was conducted across multiple metrics, including accuracy, precision, recall, and the area under the receiver operating characteristic curve (AUC). Among all classifiers, XGBoost consistently outperformed its counterparts across a range of feature combinations. Notably, the highest classification efficacy was achieved using the EMG–EEG signal pair, yielding an accuracy of 88.81\% and an AUC of 0.9747, indicating exceptional discriminative power and strong generalization to unseen data.

\section{Performance Metrics}\label{sec:Performance Metrics}
To comprehensively assess the discriminative power of the extracted statistical features correlation, cross-correlation, and covariance across signal pairs, four standard performance metrics were employed: Accuracy, Precision, Recall, and Area Under the Receiver Operating Characteristic Curve (AUC). These metrics collectively offer a multidimensional view of how effectively each classifier interprets the dynamic interactions between physiological signals for fatigue state classification \cite{32}.
\begin{figure*}[t]
    \centering
    \includegraphics[scale=0.5]{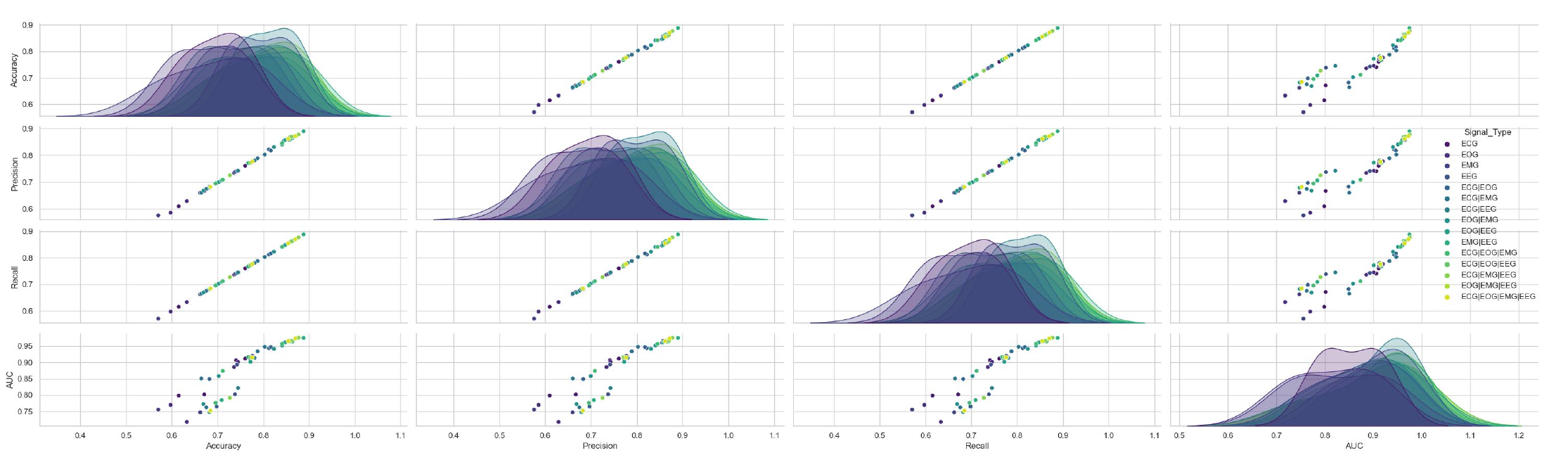}
    \caption{Pairwise scatter plot matrix of classification performance metrics across all evaluated physiological signal combinations.}
    \label{fig:PerformanceMetricsby SignalType}
\end{figure*}

\begin{itemize}
    \item Accuracy measures the proportion of correctly classified segments out of all predictions. While often treated as a baseline indicator of classifier effectiveness, its interpretability is enriched in this context by the nature of the input feature space: statistical descriptors that encode both instantaneous and lagged dependencies between modalities such as EEG, EMG, ECG, and EOG. The use of pairwise feature constructs allowed us to probe whether certain physiological relationships enhance global classification correctness across fatigue levels.
    \item Precision captures the ratio of true positive fatigue predictions to all predicted positives, and is especially informative when interpreting results from features like cross-correlation, which may emphasize transient co-activations that could result in false positives if not physiologically grounded. High precision here indicates that the classifier is effectively learning which signal interactions are consistently indicative of fatigue, rather than reacting to incidental or noisy co-fluctuations.
    \item Recall quantifies the proportion of actual fatigue cases correctly identified. It is particularly sensitive to how well the statistical signal relationships reflect underlying fatigue physiology. Features such as covariance and correlation, which capture both amplitude and directional alignment, are crucial in detecting sustained inter-signal synchrony associated with cognitive or physical fatigue. High recall demonstrates the model's ability to detect such episodes without omission, a vital requirement for safety-critical applications.
    \item AUC evaluates the classifier’s ability to discriminate between fatigue and non-fatigue classes across all possible decision thresholds. Given the subtle and often overlapping nature of physiological patterns in fatigue transitions, the AUC serves as a robust summary of separability. In this study, a one vs rest approach was applied to accommodate the multi-class structure derived from KSS-based fatigue labels. The inclusion of cross-modality statistical features provided a richer representational space, enhancing class distinction and supporting elevated AUC performance, particularly in EMG–EEG pairings.    
\end{itemize}
Together, these metrics provide a holistic evaluation framework tailored to the nature of the features explored in this study. They allow us to quantify not only the predictive performance of different classifiers but also the physiological relevance and robustness of the extracted signal relationships used as input.
\begin{table*}[t]
    \centering
    \caption{Comparative Performance of Feature Groups (Correlation, Cross-Correlation, Covariance).}
    \label{tab:ComparativePerformance}
    \begin{tabular}{l c c c c} 
        \hline
        \textbf{Feature Group} & \textbf{Best Accuracy (XGBoost)} & \textbf{Best Accuracy (RF)} & \textbf{AUC (XGBoost)} & \textbf{AUC (RF)} \\ \hline
        Correlation & 0.8881 & 0.8843 & 0.9747 & 0.9850 \\
        Cross\_correlation & 0.8823 & 0.8790 & 0.9671 & 0.9635 \\
        Covariance & 0.8810 & 0.8772 & 0.9662 & 0.9621 \\
        \hline
    \end{tabular}
\end{table*}


To further examine the relationship among the classification metrics across signal combinations, a pairwise scatter plot matrix was constructed (Figure \ref{fig:PerformanceMetricsby SignalType}). This matrix displays the joint distributions of accuracy, precision, recall, and AUC for all evaluated signal types. Diagonal elements show kernel density estimates for each metric, while off-diagonal plots reveal the linear and nonlinear interdependencies between metrics. A strong positive correlation is evident, particularly between accuracy and AUC, suggesting that high-performing models maintain consistency across multiple evaluation criteria. The data points are color-coded by signal type, revealing that combinations involving EEG, EMG, and EOG especially those in multi-modal fusion configurations tend to cluster in high-performance regions. This visual clustering confirms that feature interactions derived from cross-signal statistical descriptors contribute to robust classification. Additionally, the concentrated metric distributions for high-performing signal sets illustrate their stability across different performance dimensions. These findings support the hypothesis that modeling the statistical alignment between physiological signals can yield reliable, interpretable, and high-performing fatigue detection systems.

\section{Results}\label{sec:Results}
This study presents a systematic exploration of statistical relationships between pairs of physiological signals for fatigue detection, focusing on their effectiveness across different feature categories and machine learning models \cite{31}. The analysis provides valuable insights into how temporal coordination, amplitude co-variation, and distributional patterns between physiological signals influence classification accuracy contributing to both theoretical understanding and practical applications in biomedical signal processing, cognitive load monitoring, and wearable fatigue assessment systems.

Table \ref{tab:ComparativePerformance} summarizes the classification performance of the three principal statistical feature groups Correlation, Cross-Correlation, and Covariance across a variety of unimodal and multimodal physiological signal combinations, using two ensemble classifiers Random Forest (RF) and XGBoost. Performance evaluation was based on two key metrics, classification accuracy and area under the ROC curve (AUC), which collectively offer insight into model generalization and class separability.

The Correlation demonstrated the highest overall classification performance, with XGBoost achieving an accuracy of 88.81\% and an AUC of 0.9747, while Random Forest reached 88.43\% accuracy and 0.9688 AUC. These results highlight the discriminative power of linear interdependencies between physiological signals, particularly among EMG and EEG modalities, in distinguishing fatigue states.

Cross-Correlation, capturing temporal synchronization and delay-sensitive signal interactions, exhibited strong performance, with XGBoost and RF achieving accuracies of 88.23\% and 87.90\%, and AUC values of 0.9671 and 0.9635, respectively. These findings underscore the relevance of latency-aware dynamics and the neurophysiological coupling that arises during cognitive or physical fatigue.

The Covariance, which reflects amplitude co-variation and joint variability between signal modalities, also demonstrated high classification effectiveness. XGBoost reached 88.10\% accuracy with an AUC of 0.9662, while RF achieved 87.72\% accuracy and 0.9621 AUC. These values affirm the importance of shared amplitude modulation in capturing systemic fatigue markers, particularly when combined across multi-modal sensor inputs.

The marginal performance differences among the three statistical descriptors indicate that each contributes unique yet overlapping information. Correlation captures global co-fluctuation patterns, cross-correlation introduces temporal alignment, and covariance reflects intensity-based coupling all of which are complementary in modeling the multi-dimensional nature of fatigue.

\begin{figure}[]
    \centering
    \includegraphics[scale=0.43]{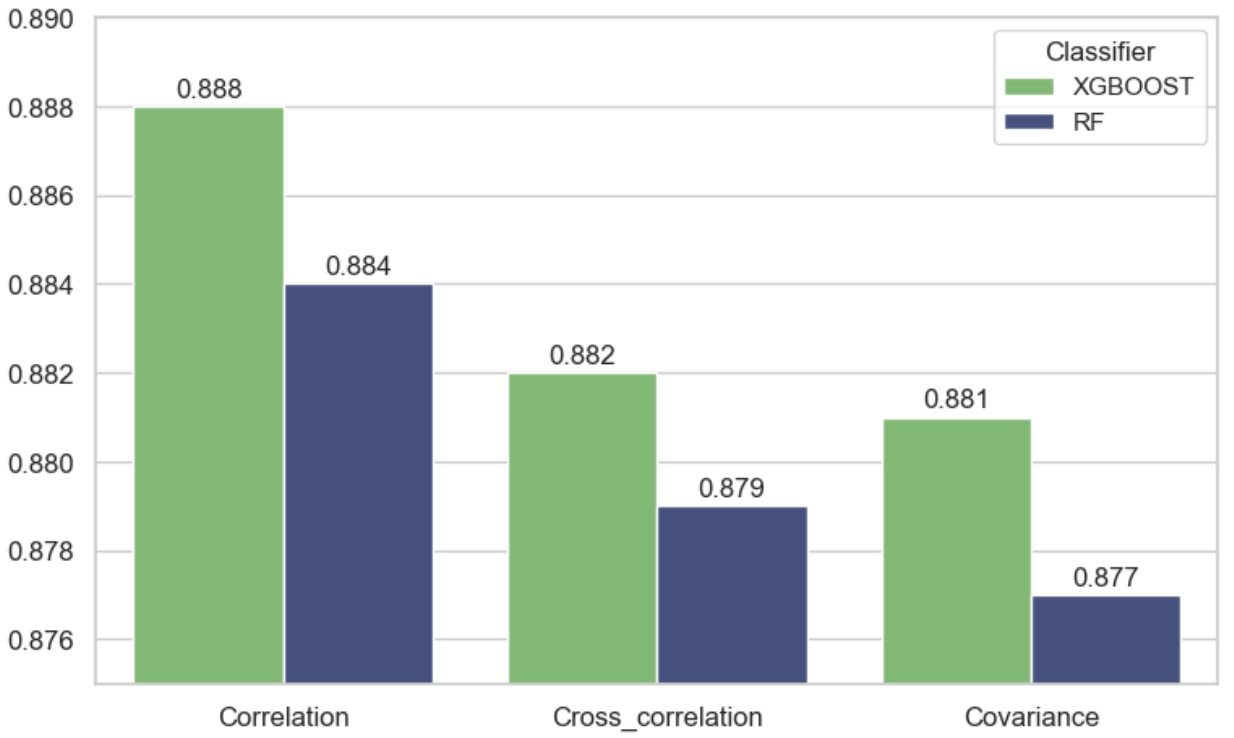}
    \caption{Accuracy comparison of XGBoost and RF across correlation, cross-correlation, and covariance features, with XGBoost showing consistently higher performance.}
    \label{fig:AccuracyPerformance}
\end{figure}

Overall, this investigation confirms that statistical relationships between physiological signals particularly those reflecting temporal alignment and amplitude co-variation are strong predictors of fatigue. The results validate the hypothesis that fatigue-induced changes manifest not only within individual signals but also in their coordinated, multi-modal interactions. This insight provides a robust foundation for building interpretable, generalizable, and scalable fatigue monitoring systems. Furthermore, the superior performance of ensemble-based classifiers especially XGBoost reinforces the value of tree-based models in learning high-dimensional signal dependencies and optimizing feature interaction modeling in complex physiological datasets.(Figure \ref{fig:AccuracyPerformance})

\subsection{Top Modalities}\label{sec:Top Modalities}
\begin{figure*}[t]
    \centering
    \includegraphics[scale=0.89]{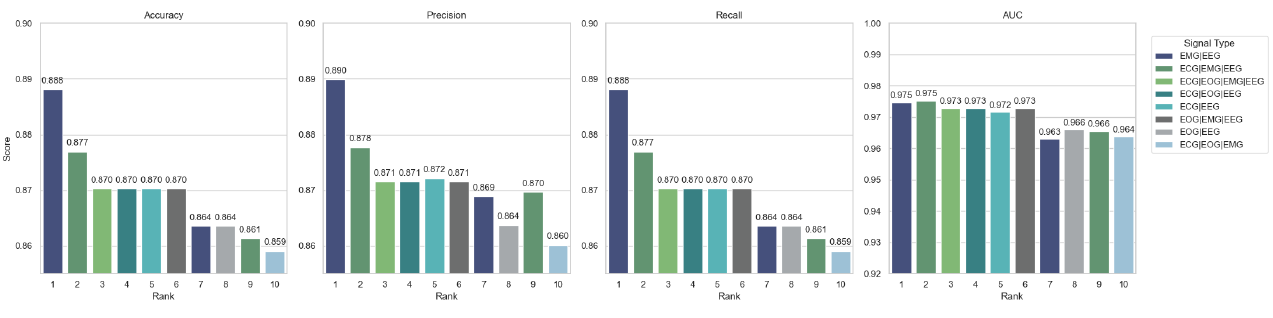}
    \caption{This chart shows the top 10 classifier setups ranked by Accuracy, Precision, Recall, and AUC.EMG|EEG consistently ranks highest across all metrics..}
    \label{fig:Top10ClassifierbyPerformaneMetrics}
\end{figure*}

Analysis of the top-performing classifiers (Table \ref{tab:Top10SignalCombinationPerformance}) reveals clear patterns in modality effectiveness. The most accurate model XGBoost applied to the EMG|EEG pairing achieved an accuracy of 88.81\% and an AUC of 0.9747, indicating that the integration of muscular and cortical activity yields the most discriminative features for fatigue classification. Additional high-ranking configurations such as ECG|EMG|EEG, ECG|EOG|EEG, and EOG|EMG|EEG also featured prominently among the top 10 classifiers, reinforcing the advantage of multi-signal fusion in capturing both autonomic and neurocognitive fatigue signatures. Notably, EMG and EEG appeared most frequently across these high-performing models, suggesting that muscular and brainwave patterns play complementary roles in reflecting fatigue-related physiological changes. In contrast, while unimodal models (e.g., EEG-only or ECG-only) offered modest predictive capability, they were consistently outperformed by their multi-modal counterparts.(Figure \ref{fig:Top10ClassifierbyPerformaneMetrics})

\begin{table*}[t]
    \centering
    \caption{Top 10 Performing Classifier-Signal Combinations Ranked by Accuracy.}
    \label{tab:Top10SignalCombinationPerformance}
    \normalsize 
    \begin{tabular}{l l c c c c} 
        \hline
        \textbf{Classifier} & \textbf{Signal Type} & \textbf{Accuracy} & \textbf{Precision} & \textbf{Recall} & \textbf{AUC} \\ 
        \hline
        \textbf{XGBOOST} & \textbf{EMG|EEG} & \textbf{0.888143} & \textbf{0.889895} & \textbf{0.888143} & \textbf{0.974705} \\
        XGBOOST & ECG|EMG|EEG & 0.876957 & 0.877669 & 0.876957 & 0.975256 \\
        XGBOOST & ECG|EOG|EMG|EEG & 0.870246 & 0.871491 & 0.870246 & 0.972853 \\
        XGBOOST & ECG|EOG|EEG & 0.870246 & 0.871491 & 0.870246 & 0.972853 \\
        XGBOOST & ECG|EEG & 0.870246 & 0.872091 & 0.870246 & 0.971586 \\
        XGBOOST & EOG|EMG|EEG & 0.870246 & 0.871491 & 0.870246 & 0.972853 \\
        RF & EMG|EEG & 0.863535 & 0.868873 & 0.863535 & 0.963033 \\
        XGBOOST & EOG|EEG & 0.863535 & 0.863650 & 0.863535 & 0.965974 \\
        RF & ECG|EMG|EEG & 0.861298 & 0.869618 & 0.861298 & 0.965574 \\
        XGBOOST & ECG|EOG|EMG & 0.859060 & 0.860091 & 0.859060 & 0.963766 \\
        \hline
    \end{tabular}
\end{table*}


Collectively, these findings demonstrate that the EMG|EEG combination offers the most effective fusion strategy for fatigue detection, delivering superior accuracy and robustness across classifiers. This suggests that while EEG remains a cornerstone in cognitive fatigue analysis, its predictive utility is greatly enhanced when integrated with EMG, rather than being used as a standalone modality. Furthermore, tri-modal combinations such as ECG|EMG|EEG and ECG|EOG|EEG yielded strong and stable classification results, supporting the broader claim that signal synergy, rather than single-channel precision, is critical for effective fatigue modeling. These insights not only validate the value of multi-modal integration but also provide a foundation for designing scalable and interpretable fatigue monitoring systems suitable for real-world deployment.

\subsection{Multi-Modal Combinations}\label{sec:Multi-Modal Combinations}
\begin{figure}[h]
    \centering
    \includegraphics[scale=0.45]{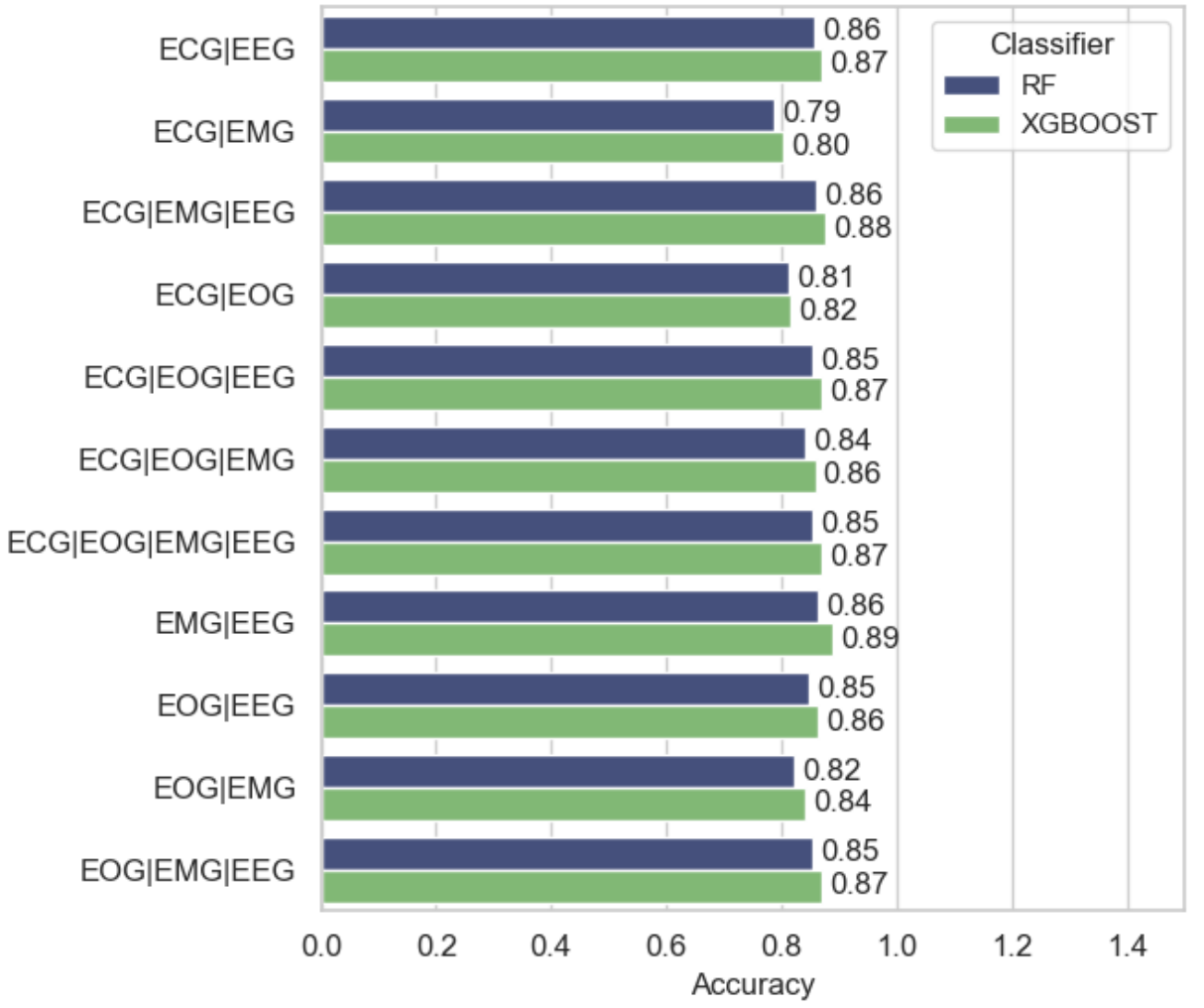}
    \caption{Comparison of classifier performance (XGBoost and Random Forest) across various signal combinations using Accuracy as the evaluation metric. Multi-modal combinations such as EMG|EEG and ECG|EMG|EEG yield the highest scores, highlighting the benefit of integrating multiple physiological signals.}
    \label{fig:MultiSignals}
\end{figure}

Figure \ref{fig:MultiSignals} illustrates the comparative accuracy of different multi-modal signal combinations using Random Forest (RF) and XGBoost classifiers. The results show that certain combinations consistently yield higher performance, underscoring the value of specific physiological modalities for fatigue detection.


The EMG | EEG combination delivers the highest performance, achieving an accuracy of 89.0\% with XGBoost and 86.0\% with RF. This pairing merges neural and neuromuscular signals and demonstrates strong discriminative power in capturing fatigue-related patterns. The results reinforce the dominance of these signals, highlighting their ability to detect both mental and physical fatigue components.

The combination of ECG | EMG | EEG follows closely, with accuracies of 88.0\% (XGBoost) and 86.0\% (RF). The inclusion of cardiovascular metrics provides additional physiological context, though the marginal performance gain over EMG | EEG suggests ECG adds only modest value in this configuration.

The full fusion scenario (ECG | EOG | EMG | EEG) achieves 87.0\% (XGBoost) and 85.0\% (RF), indicating that while comprehensive, higher-order signal integration does not significantly enhance classification accuracy compared to simpler combinations. This points to diminishing returns as more modalities are added, potentially due to overlapping or redundant information.

Tri-modal combinations such as ECG | EOG | EMG and ECG | EOG | EEG yield comparable performance (up to 87.0\% with XGBoost and 84.0--85.0\% with RF), demonstrating that ocular signals (EOG) contribute moderately when paired with ECG and either EMG or EEG. However, none of these combinations surpass the EMG | EEG benchmark.

Other configurations like EOG | EMG | EEG and EOG | EEG achieve respectable results (87.0\% and 86.0\%, respectively, for XGBoost), though slightly lower than EMG-inclusive pairs. This suggests that EOG provides useful signals, particularly in relation to drowsiness, but is not as strong a predictor as EMG or EEG alone.

Dual-modality setups such as ECG | EOG, ECG | EEG, and ECG | EMG reach accuracies ranging from 80.0\% to 87.0\%, reinforcing the idea that some physiological signals are more informative than others when fused.

These findings confirm that multi-modal signal fusion improves model accuracy but also highlight that EMG and EEG remain the most influential modalities. Simpler combinations, especially EMG | EEG, consistently outperform more complex setups, offering a balance of accuracy and implementation efficiency ideal for real-time fatigue monitoring systems.

\subsection{Performance Metrics for Multi-Modal Models}\label{sec:Performance Metrics for Multi-Modal Models}
Figure \ref{fig:Top10ClassifierbyPerformaneMetrics} presents a comparative analysis of performance metrics(accuracy, precision, recall, and AUC) for different multi-modal signal combinations evaluated using XGBoost and Random Forest (RF) classifiers. This multifaceted evaluation provides a holistic understanding of how each classifier handles variations in physiological data for fatigue detection.

The \textit{EMG | EEG} combination consistently ranks highest across all four metrics, with XGBoost achieving 89.0\% accuracy, 89.0\% precision, 89.0\% recall, and an AUC of 0.975. RF follows closely with 86.0\% across accuracy, precision, and recall, and an AUC of 0.966. These results underscore the synergy between neuromuscular and neural signals in detecting fatigue, supporting their combined use for capturing both physical and cognitive fatigue characteristics.

\textit{ECG | EMG | EEG} and \textit{ECG | EOG | EMG} also demonstrate strong and balanced performance across all metrics, with XGBoost achieving up to 88.0\% accuracy and AUC values nearing 0.974. These tri-modal configurations highlight the value of integrating autonomic and muscular indicators to strengthen classifier robustness, especially for real-world applications requiring both sensitivity and generalizability.

Across classifiers, XGBoost consistently outperforms RF by small but consistent margins across all signal types. Notably, precision and recall remain closely aligned for both classifiers, suggesting minimal trade-offs between false positives and false negatives. This consistency further supports the application of these models in safety-critical domains, such as driver monitoring or industrial fatigue detection.

Higher-order signal fusions like \textit{ECG | EOG | EMG | EEG} show strong AUC values but offer limited improvements in accuracy or precision compared to simpler tri-modal or dual-modal models. This suggests diminishing returns when adding more physiological modalities, potentially due to redundancy or noise introduced by overlapping signal domains.

EEG and EOG combinations alone or without EMG tend to yield lower precision and recall values, despite achieving moderate AUC scores. This highlights the limited standalone discriminatory power of brain and ocular signals for fatigue classification. However, their integration with EMG or ECG significantly boosts their utility, particularly for scenarios emphasizing cognitive or drowsiness-related fatigue.

In summary, the performance metrics affirm that multi-modal combinations(particularly those including EMG and ECG) enhance fatigue classification accuracy while maintaining balanced recall and precision. Simpler dual- and tri-modal combinations remain optimal for implementation, offering both high classification reliability and model efficiency across real-time fatigue detection use cases.

\subsection{Technical Discussion on Classifier Trends}\label{sec:Technical Discussion on Classifier Trends}
The classifier performance trends in Table \ref{tab:Top10SignalCombinationPerformance} and Figure \ref{fig:Top10ClassifierbyPerformaneMetrics} underscore several technical patterns in the behavior of multi-modal models for fatigue detection. Notably, XGBoost consistently outperforms Random Forest (RF) across all key metrics especially when applied to high-order signal combinations that capture diverse physiological characteristics.

The \textit{EMG | EEG} combination emerges as the most effective setup, with XGBoost achieving the highest recorded accuracy (88.8\%) and AUC (0.9747). Its balanced precision and recall scores (both 88.9\%) indicate that this configuration excels in identifying both fatigued and non-fatigued states with minimal trade-offs. The inclusion of EMG highlights its role in detecting muscular fatigue, while EEG contributes cognitive fatigue insights, making their fusion highly complementary.

Combinations involving three or more signals (such as \textit{ECG | EMG | EEG} and \textit{ECG | EOG | EMG}) also demonstrate high accuracy (87.0--87.6\%) and robust AUC scores (up to 0.9752). These setups integrate autonomic, ocular, and neuromuscular features, enabling classifiers to detect fatigue with high granularity across physical and mental domains. However, the marginal performance improvement over simpler pairs like EMG | EEG suggests that additional signals introduce slight gains in robustness at the cost of increased model complexity.

Interestingly, the full fusion model \textit{ECG | EOG | EMG | EEG} ranks fifth in accuracy (87.0\%) and shares the same AUC (0.9728) as some tri-modal setups, suggesting diminishing returns when excessive modalities are combined. This may stem from overlapping signal contributions or increased noise that dilutes the discriminative power of individual modalities.
Dual-modality combinations such as \textit{EOG | EEG} and \textit{EOG | EMG | EEG} rank lower in both accuracy (85.0--86.3\%) and AUC (0.9659--0.9708), reinforcing that ocular and neural signals, while informative, lack standalone discriminative strength without muscular or autonomic support. The model \textit{ECG | EOG | EEG} performs slightly better than these, but still falls behind the top tri-modal combinations.

From a classifier standpoint, XGBoost consistently provides superior performance across all combinations, even in subtle metrics like recall and precision alignment. Its gradient boosting structure is likely more effective in modeling non-linear relationships between features extracted from diverse physiological domains. RF, while slightly behind in metrics, offers comparable stability and may be better suited for real-time or resource-constrained deployments due to its simpler tree-based ensemble architecture.

Overall, the observed classifier trends validate the hypothesis that multi-modal fusion(particularly combinations involving EMG and either EEG or ECG) yields high-performing, reliable models for fatigue detection. They also affirm the utility of ensemble classifiers like XGBoost in managing high-dimensional, multi-source physiological data while maintaining robust generalization across fatigue-related classification tasks.

\section{SHAP-Based Model Explainability Analysis}\label{sec:SHAP-Based Model Explainability Analysis}
To deepen the interpretability of the proposed fatigue detection system, we applied SHAP (SHapley Additive exPlanations) to the best-performing configuration, the XGBoost classifier trained on the EMG | EEG signal combination. This configuration achieved the highest performance across all tested models, with Accuracy = 0.8881, Precision = 0.8890, Recall = 0.8881, and AUC = 0.9747 (Table \ref{tab:Top10SignalCombinationPerformance}). Given its superior discriminative capability, EMG | EEG was selected as the primary target for feature attribution and interpretability analysis.

SHAP provides a unified approach for quantifying each feature’s contribution to a model’s prediction, decomposing the difference between the baseline output (expected value) and the actual prediction into additive feature attributions. This approach supports global interpretability, identifying consistently important features across the dataset, and local interpretability explaining how feature values for individual samples drive classification outcomes.
\subsection{Global Feature Importance and Distribution Patterns}\label{sec:Global Feature Importance and Distribution Patterns}
\begin{figure}[h]
    \centering
    \includegraphics[scale=0.40]{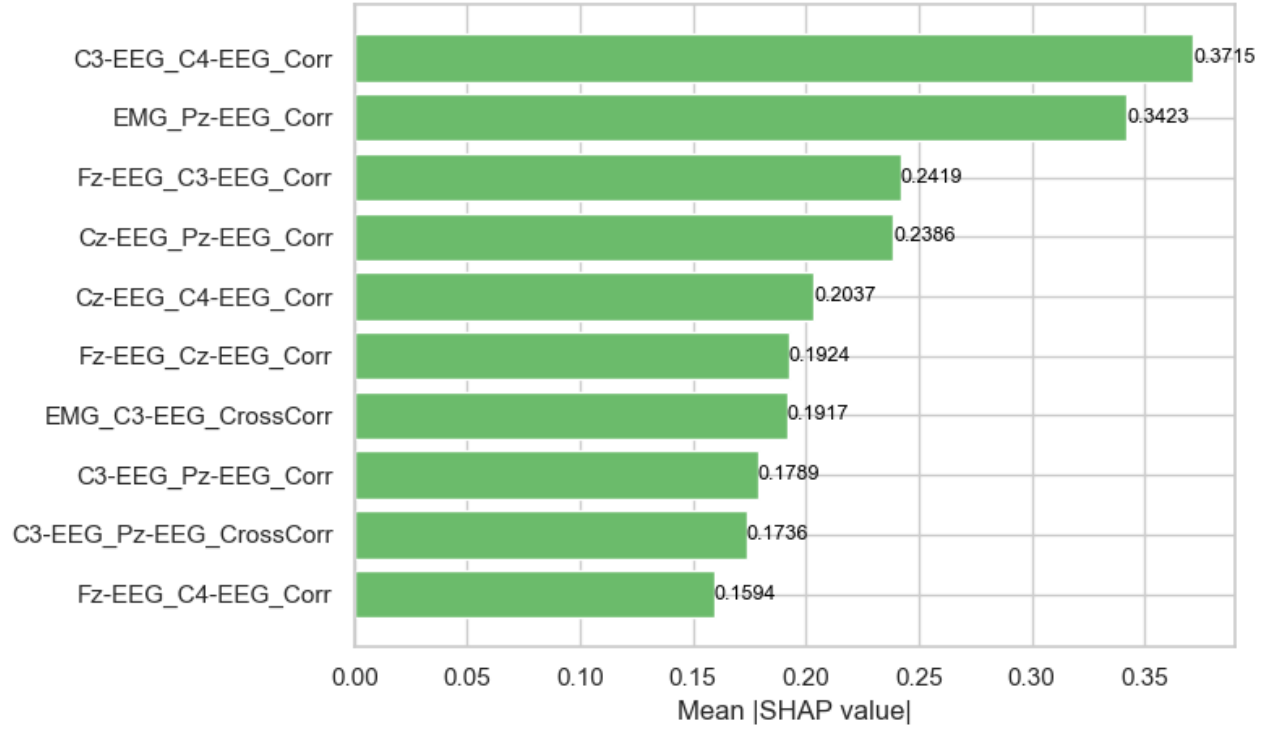}
    \caption{Top-10 feature importance for the Fatigued class from the EMG|EEG model, ranked by mean absolute SHAP values.}
    \label{fig:ShapTop10Features}
\end{figure}

The global feature ranking for the Fatigued class, obtained from mean absolute SHAP values, is shown in Figure \ref{fig:ShapTop10Features}. The most influential predictors were EEG--EEG correlation features such as C3--EEG\_C4--EEG\_Corr and Fz--EEG\_C3--EEG\_Corr, along with EMG--EEG correlations such as EMG\_Pz--EEG\_Corr. Negative contributions were predominantly associated with EMG--EEG cross-correlation features, for example, EMG\_Cz--EEG\_CrossCorr.

The analysis reveals how variations in feature values correspond to positive or negative shifts in model predictions, highlighting both the direction and magnitude of their influence.

\begin{figure}[h]
    \centering
    \includegraphics[scale=0.40]{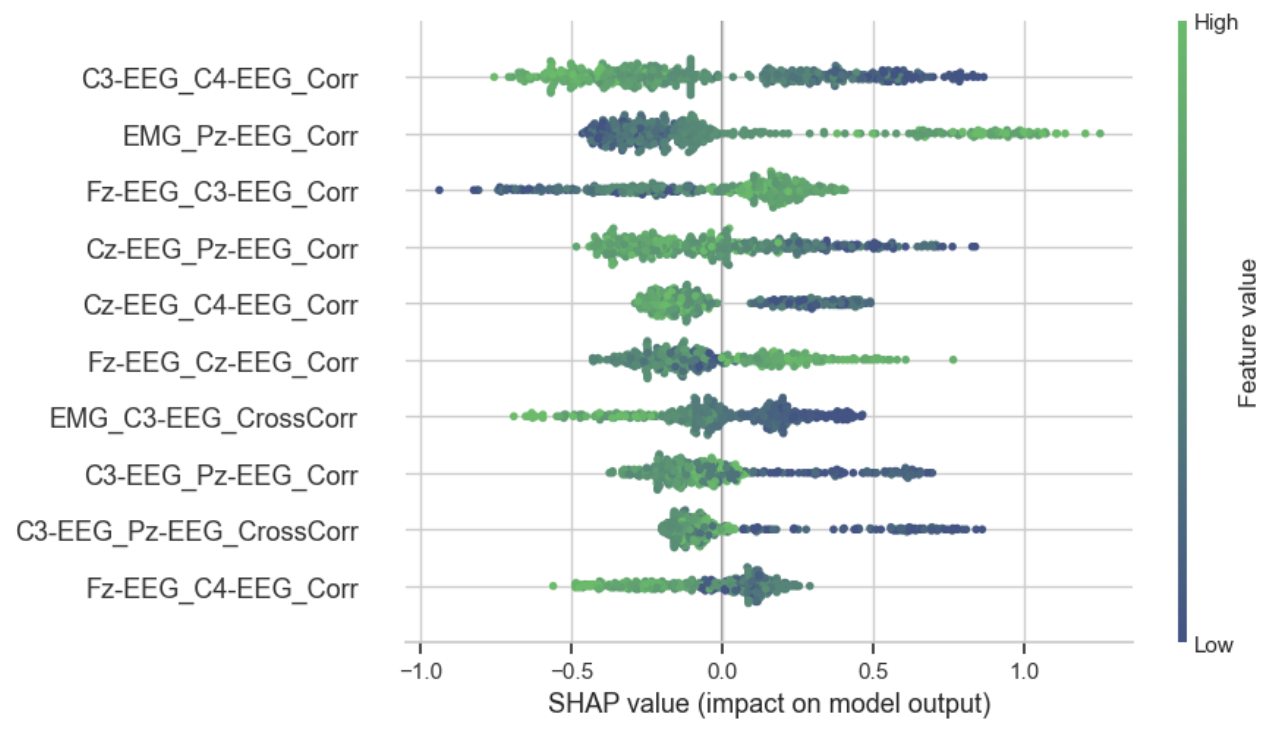}
    \caption{Top 10 SHAP-ranked features for the Fatigued class in the EMG | EEG XGBoost model, showing feature value (color) and impact on prediction.}
    \label{fig:ShapFatiguedTop10}
\end{figure}

\begin{figure}[h]
    \centering
    \includegraphics[scale=0.40]{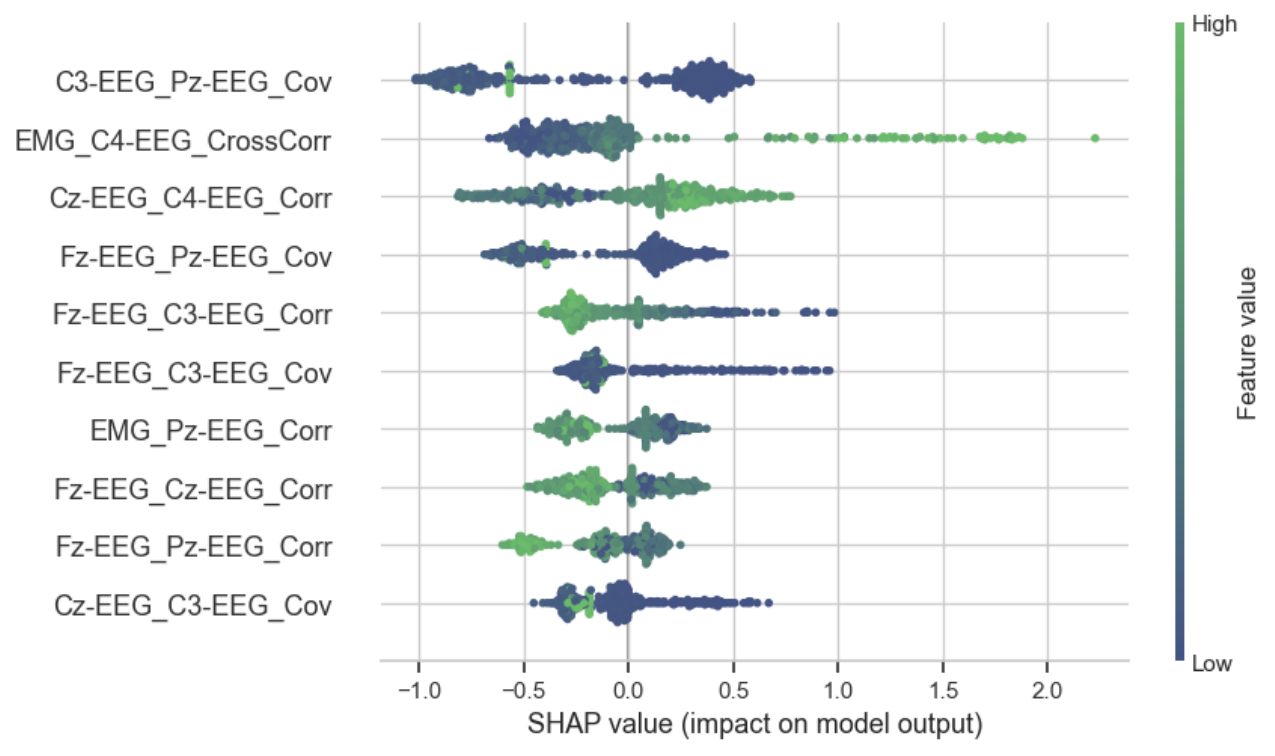}
    \caption{Top 10 SHAP-ranked features for the Normal class in the EMG | EEG XGBoost model, with color indicating feature value and horizontal spread showing impact on prediction.}
    \label{fig:ShapNormalTop10}
\end{figure}

In the Fatigued class (Figure \ref{fig:ShapFatiguedTop10}), elevated values of C3--EEG\_C4--EEG\_Corr are strongly associated with increased model propensity toward fatigue classification, whereas specific EMG--EEG cross-correlation patterns exert a counteracting effect, reducing the likelihood of a fatigued prediction. The Normal class (Figure \ref{fig:ShapNormalTop10}) exhibits a more evenly distributed influence of EEG--EEG and EMG--EEG features, indicative of broader and less sharply defined decision boundaries. 

\begin{figure}[h]
    \centering
    \includegraphics[scale=0.40]{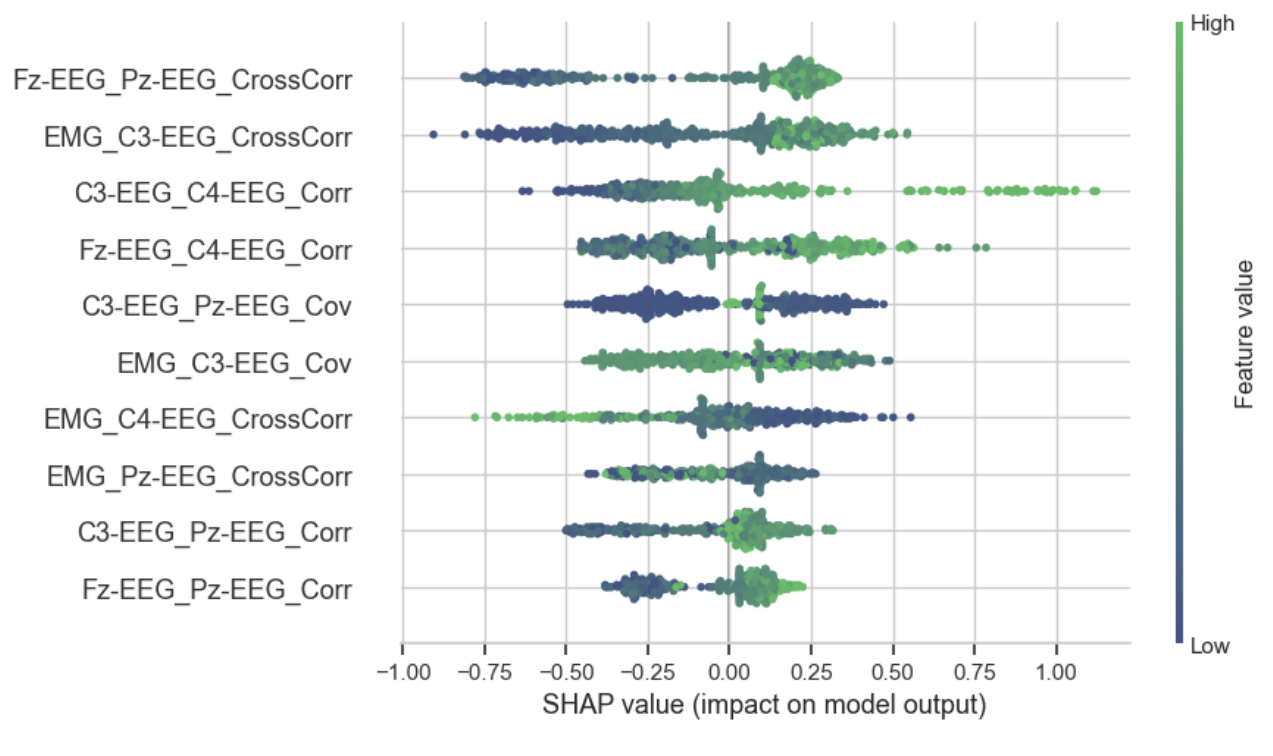}
    \caption{Top 10 SHAP-ranked features for the Alert class in the EMG | EEG XGBoost model, illustrating the influence of each feature value on prediction outcomes.}
    \label{fig:ShapAlertTop10}
\end{figure}

In contrast, the Alert class (Figure \ref{fig:ShapAlertTop10}) is primarily distinguished by strong connectivity patterns between central and parietal EEG channels, while EMG-related features play only a minor role in the model’s decision-making process. These findings confirm the physiological plausibility of the learned patterns, aligning with established neuro-muscular--cortical dynamics observed during fatigue states.

\subsection{Localized Instance-Level Explanations}\label{sec:Localized Instance-Level Explanations}
\begin{figure*}[t]
    \centering
    \includegraphics[scale=0.85]{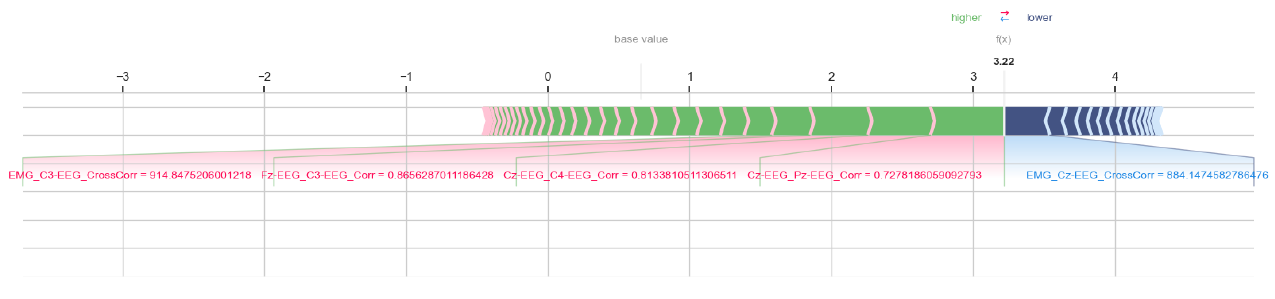}
    \caption{SHAP force plot for a representative fatigued instance, illustrating how key physiological features collectively influence the model’s output toward a strong fatigue classification.}
    \label{fig:ShapForcePlot}
\end{figure*}

To explore how the model arrives at a prediction for an individual case, a SHAP force plot (Figure \ref{fig:ShapForcePlot}) was generated for a Fatigued instance. The visualization decomposes the final prediction into the baseline plus additive contributions from each feature, with green arrows increasing and blue arrows decreasing the predicted probability of fatigue.

The force plot reveals that for this sample, strong positive contributions came from C3\allowbreak--EEG\_\allowbreak C4\allowbreak--EEG\_\allowbreak Corr, Fz\allowbreak--EEG\_\allowbreak C3\allowbreak--EEG\_\allowbreak Corr, and EMG\_\allowbreak Pz\allowbreak--EEG\_\allowbreak Corr, whereas EMG\_\allowbreak Cz\allowbreak--EEG\_\allowbreak CrossCorr had a strong negative contribution, reducing the fatigue score. This localized perspective confirms the global patterns while offering case-specific interpretability for decision support.

\subsection{Feature Contribution Analysis}\label{sec:Feature Contribution Analysis}
\begin{figure}[h]
    \centering
    \includegraphics[scale=0.40]{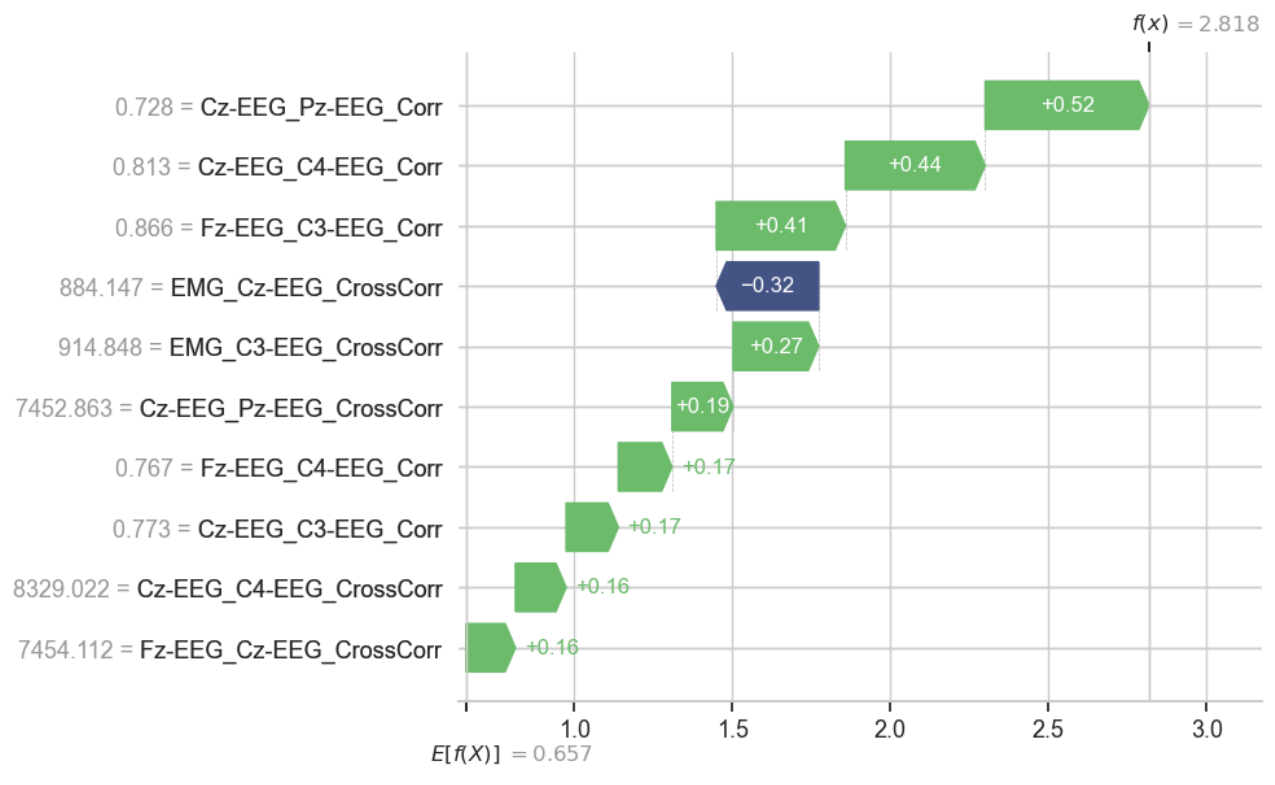}
    \caption{Waterfall plot of the top ten features for a sample instance, showing positive EEG correlations driving the prediction upward and EMG|EEG cross-correlations reducing it.}
    \label{fig:ShapWaterFallTop10}
\end{figure}

The waterfall plot (Figure \ref{fig:ShapWaterFallTop10}) for the same instance provides a stepwise decomposition of the model's output into the contributions of the top ten most influential features, arranged in descending order of absolute impact. This visualization traces the cumulative progression from the model's baseline expectation \( E[f(X)] \) (representing the average predicted value across the training dataset) to the specific prediction for the instance under consideration.

Within this sample, positive contributions are predominantly driven by correlations between cortical EEG channels, which collectively shift the prediction toward a higher fatigue score. In contrast, certain EMG|EEG cross-correlation features act as negative contributors, exerting an opposing influence that counterbalances the upward shift induced by the EEG-related features.

To ensure interpretability and reduce visual complexity, the plot has been restricted to the ten most influential features, thereby omitting the aggregate “other features” category. This focused representation enables a clearer examination of the most critical predictors, facilitating both domain-specific interpretation and cross-validation of the learned model behavior.
\subsection{Cross-Class Comparisons}\label{sec:Cross-Class Comparisons}
Class-specific SHAP analyses highlight clear differences in the distribution and magnitude of feature importance across the three alertness states, reflecting the model’s ability to capture distinct neuromuscular–cortical signatures. In the Fatigued class (Figure \ref{fig:ShapFatiguedTop10}), the predictive signal is primarily driven by strong frontal–central EEG correlations, complemented by a subset of EMG|EEG cross-correlation features, indicating heightened integration between cortical and muscular activity during fatigue onset. In contrast, the Normal class (Figure \ref{fig:ShapNormalTop10}) shows a more balanced distribution of contributing factors, indicative of a broader and less sharply defined decision boundary, where no single source of information exerts dominant influence over the classification outcome. For the Alert class (Figure \ref{fig:ShapAlertTop10}), the most influential predictors shift toward central–parietal EEG correlations, with minimal contribution from EMG-related features, implying a reduced neuromuscular component and greater reliance on cortical synchrony in well-rested states. Collectively, these differences demonstrate that the model effectively differentiates alertness states by leveraging physiologically plausible and state-specific neural–muscular interaction patterns.

\subsection{Summary of Insights from SHAP Analysis}\label{sec:Summary of Insights from SHAP Analysis}
The SHAP-based evaluation of the EMG | EEG XGBoost model provided a robust interpretability framework that yielded several important conclusions regarding the model’s decision-making processes and their physiological plausibility. First, the analysis confirmed that the most predictive indicators of fatigue arise from correlation and cross-correlation metrics involving frontal and central regions, reflecting dynamic interactions between cortical activity and muscular responses that are well-documented in fatigue research. Second, the directionality of feature contributions revealed a consistent pattern in which high correlation values were associated with increased probability of fatigue detection, whereas specific interaction patterns exerted a negative influence, attenuating the likelihood of a fatigued classification. Third, the combined use of global SHAP summaries and local instance-level explanations enhanced transparency by offering both an overarching view of feature importance and a fine-grained account of how individual predictions are formed, ensuring reproducibility and model accountability. Finally, the recurrence of physiologically meaningful features across multiple instances supports the biological validity of the model’s learned representations, reinforcing confidence that its predictive performance is grounded in genuine neuromuscular–cortical dynamics rather than spurious statistical correlations.

\section{Research Value and Impact}\label{sec:Research Value and Impact}
This study systematically investigated the role of statistical relationships specifically correlation, cross-correlation, and covariance between pairs of physiological signals in improving fatigue classification. Our findings demonstrate that modeling inter-signal dependencies provides a rich, low-complexity, and highly discriminative feature space for machine learning classifiers. The superior performance of XGBoost using the EMG|EEG combination (accuracy: 88.81\%, AUC: 0.975) emphasizes the value of signal fusion in enhancing classifier robustness and generalizability. 

\subsection{Interpretation of Key Findings}\label{sec:Interpretation of Key Findings}
The present study revealed that the combination of EMG and EEG signals using XGBoost resulted in the highest performance across all statistical feature groups. Specifically, the Correlation feature group yielded an accuracy of 88.81\% and an AUC of 0.9747. These findings underscore the informativeness of linear relationships between physiological signals, particularly those related to muscular and neural activity, in modeling fatigue. This is consistent with prior research emphasizing the discriminative potential of EEG and EMG signals in fatigue detection contexts \cite{4}, \cite{5}, \cite{13}, \cite{14}.

Moreover, the use of XGBoost across all feature groups consistently outperformed Random Forest, demonstrating its robustness and adaptability to complex, high-dimensional feature spaces. The ability of XGBoost to capture non-linear interactions while maintaining computational efficiency adds practical value for near real-time or embedded fatigue detection systems \cite{13}, \cite{25}.
\subsection{Comparison with Existing Studies}\label{sec:Comparison with Existing Studies}
Compared to deep learning approaches such as CNNs \cite{4}, \cite{18}, attention-based CNNs \cite{20}, LSTM models \cite{33}, and temporal convolutional networks (TCNs) \cite{34}, our approach achieves comparable or superior performance while maintaining significantly lower model complexity and improved interpretability. For example, Lin et al. \cite{4} used CNNs with transfer learning on EEG data and achieved 87\% accuracy, while our EMG|EEG Correlation model with XGBoost achieved 88.81\%. Similarly, Yang et al. \cite{18} achieved 87.5\% accuracy using EEG spectral features and CNNs, but without model explainability.

Studies such as Ma et al. \cite{14} and Hou et al. \cite{13} demonstrated the effectiveness of multimodal physiological signals for fatigue detection using joint signal representations or XGBoost classifiers. Our results align with these findings but extend the value by incorporating transparent statistical features and SHAP-based interpretability, which is rarely addressed in the context of EMG|EEG fusion.

Furthermore, the superior performance of our correlation-based features complements the findings of Kim et al. \cite{23}, who reported that cross-correlation and correlation metrics were informative in multimodal settings. However, their work lacked a model-agnostic explanation framework like SHAP, which we incorporated for both local and global interpretability \cite{25}.

The robustness of our EMG|EEG combination outperformed traditional unimodal approaches such as ECG-only models evaluated by Liu et al. \cite{21} and Li et al. \cite{5}, suggesting that signal diversity from different physiological systems is beneficial. Finally, the cross-subject generalization challenge addressed by Xu et al. \cite{35} using transfer learning remains complementary to our approach, which is grounded in robust feature engineering and ensemble learning but may benefit from future integration of TL methods.
\subsection{Contributions to the Field}\label{sec:Contributions to the Field}
This study introduces a transparent and physiologically grounded fatigue detection framework based on statistical feature extraction and SHAP-based model explainability. Compared to black-box deep learning models \cite{18}, \cite{33}, our method offers clear advantages in interpretability, reproducibility, and clinical trust, in line with the goals of explainable AI in healthcare \cite{9}, \cite{25}.

The high performance achieved by simple statistical features challenges the notion that fatigue detection must rely on computationally intensive deep learning architectures. Our results encourage the use of domain-informed feature engineering coupled with interpretable models as a viable pathway for real-world fatigue monitoring systems \cite{1}, \cite{2}, \cite{13}, \cite{14}, \cite{31}.

\subsection{Limitations and Considerations}\label{sec:Limitations and Considerations}
While our results are promising, generalizability may still be influenced by participant variability, signal quality, and sensor placement. The dataset (DROZY) offers a controlled experimental setting, but cross-domain application (e.g., driving, industrial monitoring) would benefit from further validation. Our study did not use deep temporal modeling (e.g., LSTM or TCNs), which may capture sequential fatigue trends more effectively \cite{33}, \cite{34}. However, this was a deliberate tradeoff to retain model simplicity and explainability.

\subsection{Practical Implications and Future Applications}\label{sec:Practical Implications and Future Applications}
The implications of our findings are substantial. Accurate and interpretable fatigue detection has critical applications in driving safety, workplace productivity, and clinical diagnostics \cite{1}, \cite{2}, \cite{5}, \cite{13}, \cite{21}. Our proposed framework can serve as the basis for developing lightweight wearable systems that provide real-time fatigue feedback with clear reasoning for model predictions.

By integrating biologically meaningful signals (EMG and EEG) and emphasizing transparency, this research lays the groundwork for fatigue detection systems that are not only performant but also trustworthy and actionable a necessity for adoption in health-sensitive environments \cite{9}, \cite{25}.

\section{Final Remarks and Future Work}\label{sec:Final Remarks and Future Work}	
This study introduced a robust, interpretable framework for fatigue detection by leveraging statistical relationships(specifically correlation, covariance, and cross-correlation) across multimodal physiological signals. By systematically evaluating combinations of EEG, ECG, EOG, and EMG signals, we demonstrated that simple yet informative features, when combined with classical machine learning classifiers such as XGBoost and Random Forest, can yield highly competitive performance. Notably, the EMG|EEG combination achieved the highest classification accuracy (0.888) and AUC (0.975), highlighting its potential as a compact and effective signal set for fatigue monitoring.

SHAP-based explainability analyses further strengthened the biological and technical interpretability of our findings, offering insights into signal synergy and model decision logic. These insights are particularly valuable for translating fatigue detection systems into real-world applications, where transparency and reproducibility are critical.

Future work will focus on several directions to extend this foundation:
\begin{itemize}
    \item Feature Expansion: Incorporating higher-order statistical descriptors (e.g., entropy, mutual information, nonlinear coupling) to capture more nuanced physiological dynamics.

    \item Time-Series Modelling: Adopting sequential models such as LSTM \cite{33} or temporal convolutional networks (TCNs) \cite{34} to account for temporal dependencies and transitions between fatigue levels.

    \item Transfer Learning: Applying transfer learning techniques across subjects or datasets to improve generalizability and reduce the need for subject-specific calibration \cite{35}.

    \item Wearable Integration: Testing the selected signal combinations on real-time wearable devices to evaluate computational feasibility and sensor placement practicality.

    \item Multi-Class \& Real-Time Systems: Extending the binary classification approach to multi-class fatigue grading and implementing real-time fatigue alert systems with adaptive thresholds.

\end{itemize}

Overall, the proposed framework serves as a foundation for low-complexity, explainable, and multi-sensor fatigue assessment, paving the way for broader adoption in domains such as occupational health, cognitive performance monitoring, and automotive safety.

\bibliographystyle{elsarticle-num} 
\bibliography{main.bib}

\end{document}